\documentclass[letterpaper]{article} 
\usepackage{aaai25}  
\usepackage{times}  
\usepackage{helvet}  
\usepackage{courier}  
\usepackage[hyphens]{url}  
\usepackage{graphicx} 
\usepackage{natbib}  
\usepackage{caption} 
\usepackage{algorithm}
\usepackage{algorithmic}

\newcommand{\decrease}[1]{
	{\fontsize{7pt}{0.5em}\selectfont\color{red!100}{$\downarrow$~{#1}}}
}
\usepackage{newfloat}
\usepackage{listings}
\usepackage{amsmath}
\usepackage{booktabs}
\usepackage{colortbl}
\usepackage{xcolor}
\usepackage{amsmath}
\usepackage{multirow}

\usepackage{newfloat}
\usepackage{listings}
\DeclareCaptionStyle{ruled}{labelfont=normalfont,labelsep=colon,strut=off} 
\floatstyle{ruled}
\newfloat{listing}{tb}{lst}{}
\floatname{listing}{Listing}
\title{ST$^3$: Accelerating Multimodal Large Language Model by
\\ Spatial-Temporal Visual Token Trimming}
\author {
    Jiedong Zhuang\textsuperscript{\rm 1,2},
    Lu Lu\textsuperscript{\rm 2},
    Ming Dai\textsuperscript{\rm 1},
    Rui Hu\textsuperscript{\rm 1},
    Jian Chen\textsuperscript{\rm 2},
    Qiang Liu\textsuperscript{\rm 2},
    Haoji Hu\textsuperscript{\rm 1}\corresponding
}
\affiliations {
    \textsuperscript{\rm 1}Zhejiang University\\
    \textsuperscript{\rm 2}Alibaba Cloud Computing\\
    zhuangjiedong@zju.edu.cn, ll200214@alibaba-inc.com, mingdai997@outlook.com, rickhu@zju.edu.cn, j.chen@alibaba-inc.com,
    yf.yifeng@alibaba-inc.com, haoji\_hu@zju.edu.cn
}

\usepackage{bibentry}

\begin{document}

\maketitle

\begin{abstract}
Multimodal large language models (MLLMs) enhance their perceptual capabilities by integrating visual and textual information. However, processing the massive number of visual tokens incurs a significant computational cost. 
Existing analysis of the MLLM attention mechanisms remains shallow, leading to coarse-grain token pruning strategies that fail to effectively balance speed and accuracy. 
In this paper, we conduct a comprehensive investigation of MLLM attention mechanisms with LLaVA. We find that numerous visual tokens and partial attention computations are redundant during the decoding process. Based on this insight, we propose Spatial-Temporal Visual Token Trimming ($\textbf{ST}^{3}$), a framework designed to accelerate MLLM inference without retraining. $\textbf{ST}^{3}$ consists of two primary components: 1) Progressive Visual Token Pruning (\textbf{PVTP}), which eliminates inattentive visual tokens across layers, and 2)  Visual Token Annealing (\textbf{VTA}), which dynamically reduces the number of visual tokens in each layer as the generated tokens grow. Together, these techniques deliver around $\mathbf{2\times}$ faster inference with only about $\mathbf{30\%}$ KV cache memory compared to the original LLaVA, while maintaining consistent performance across various datasets. 
Crucially, $\textbf{ST}^{3}$ can be seamlessly integrated into existing pre-trained MLLMs, providing a plug-and-play solution for efficient inference. 
\end{abstract}

%

\section{Introduction}
The field of generative AI has recently witnessed explosive advancements, primarily driven by the development of large language models (LLMs)~\cite{gpt4,llama2}.
Building upon this progress, Multimodal Large Language Models (MLLMs)~\cite{llava,llava1.5,llava-next} have emerged as a powerful approach, integrating vision encoders like CLIP~\cite{CLIP} to extract visual features. 
This integration enhances the reasoning capabilities of LLMs, enabling them to excel in complex tasks such as visual question answering (VQA)~\cite{SQA,ai2d}, visual reasoning~\cite{mme,mmmu,mmbench} and visual grounding~\cite{simvg,falip}.
However, the substantial computational demands of MLLMs pose a significant challenge, with their billions of parameters demanding exponential increases in computational resources as the input sequence length grows. 

\begin{figure}[t]
\centering
\includegraphics[width=0.45\textwidth]{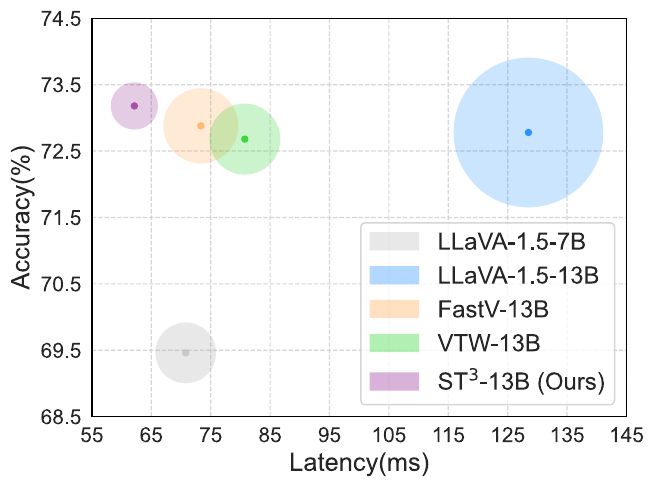}
\caption{Comparison of various models on dataset ScienceQA\_Img~\cite{SQA} with the circle size representing their FLOPs. Our method achieves the highest accuracy in 13B parameter models while maintaining the minimum FLOPs and decoding latency, even outperforming the smaller model LLaVA-1.5-7B.}
\label{fig_one}
\end{figure}

\begin{figure*}[t]
\centering
\includegraphics[width=1\textwidth]{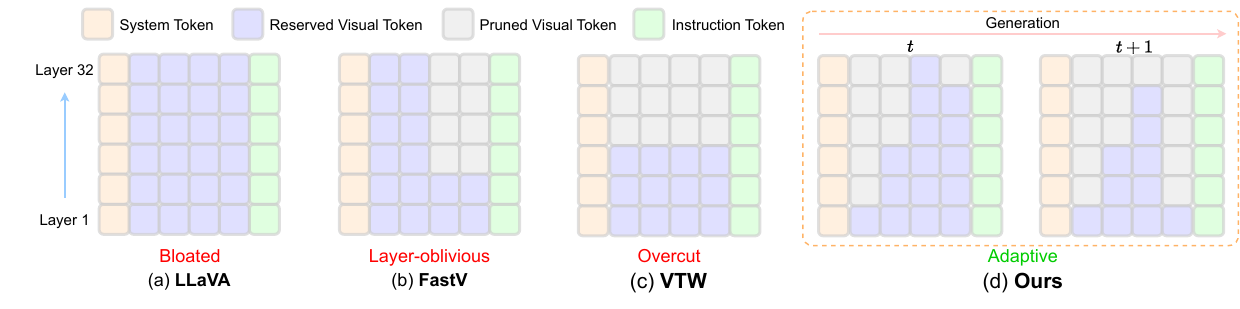}
\caption{
Illustration of our method compared with existing visual token pruning methods. (a) LLaVA~\cite{llava1.5} keeps all visual tokens. (b) FastV~\cite{fastv} prunes a fixed number of tokens in deeper layers. \textit{This layer-oblivious paradigm overlooks the variability of attention patterns across layers}. (c) VTW~\cite{VTW} prunes all visual tokens in the latter half layers, \textit{leading to a permanent loss of visual information in deeper layers}. Additionally, these three methods maintain the same quantity of visual tokens throughout the entire generation process, \textit{requiring a substantial KV cache memory budget}. (d) Our method prunes inattentive visual tokens progressively as the layer goes deeper, while dynamically reducing tokens in the generation process. It maximizes the inference efficiency by exploiting the limit of the model's dependence on visual tokens.}
\label{fig_token_prune_compare}
\end{figure*}

Moreover, converting high-resolution images into visual tokens further exacerbates the inference cost~\cite{llava-next,internlm4k,feast}, creating a significant obstacle to the practical applications of MLLMs in real-time scenarios.
Previous methods~\cite{honeybee,blip,feast,SeTok} attempt to address this by {\it introducing learnable modules} before feeding the tokens into the model, aiming to {\it reduce the length of the visual sequence}. 
While these methods can effectively reduce the inference latency, they typically require fine-tuning or retraining the entire MLLM. 
Given the current trend of rapidly increasing model sizes, the requirement of retraining or fine-tuning is suboptimal, since these processes are extremely resource-intensive and time-consuming for models with billions of parameters. 
On the other hand, recent works~\cite{fastv,VTW} have focused on reducing the number of visual tokens in the inference process without retraining.
These studies analyzed the attention scores within the decoder and observed that the ``Attention Sink'' phenomenon~\cite{streamllm} is also present in MLLMs: a majority of the visual tokens receive a low level of attention, and these {\it unattended tokens can be directly pruned with negligible impact} on the model's overall performance.
However, these methods~\cite{fastv,VTW} overlook two important aspects: 1) the iterative nature of the text generation process, where each newly generated token affects the allocation of visual attention in subsequent generation steps; 2) the potential performance gain from leveraging the internal distribution of the collection of visual tokens. 

\begin{figure*}[t]
\centering
\includegraphics[width=1\textwidth]{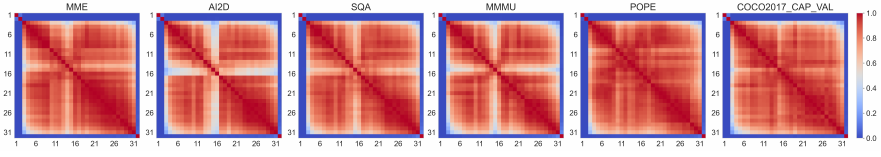}
\caption{Illustration of the similarity between the attention scores of all layers in LLaVA-1.5-7B. High similarity scores are distributed around the diagonal, indicating adjacent layers have more similar attention patterns.}
\label{fig_cos_qk}
\end{figure*}

\begin{figure}[t]
\centering
\includegraphics[width=0.47\textwidth]{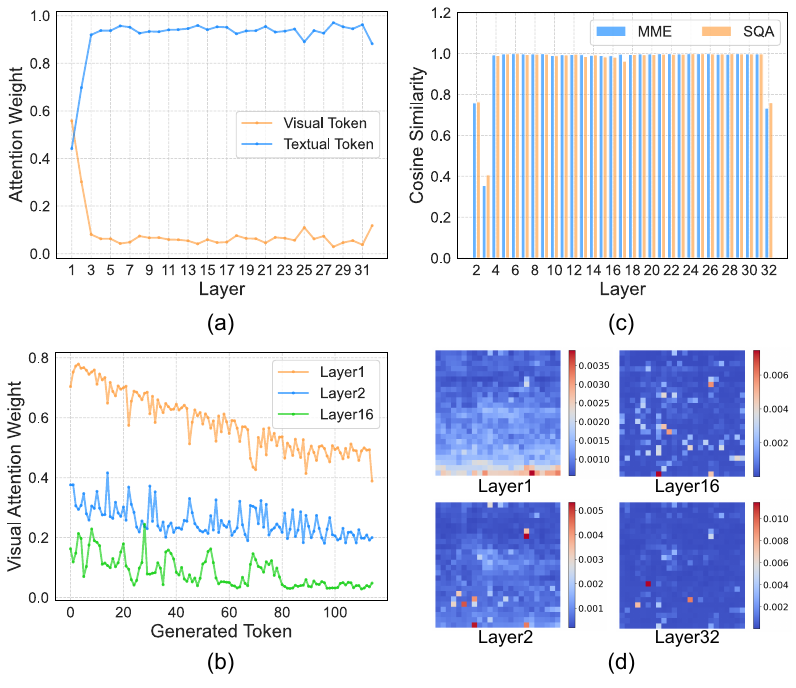}
\caption{(a) Attention weight in various layers. Visual attention exhibits a persistently low magnitude after the layer3. (b) The weight of visual attention changes with the length of the generated text sequence. (c) Cosine similarity of the attention scores between each layer and its previous layer. A value close to 1 indicates that the attention score distributions in the two layers are nearly identical. (d)Visual attention. High attention tokens decrease as layers deepen.}
\label{fig_2}
\end{figure}

Although some token pruning methods~\cite{streamllm,h2o,snapkv} have been proposed in LLM, token pruning in MLLM remains relatively underexplored.
Due to the inherent differences between the modalities of images and texts, it is challenging to directly apply existing LLM-based techniques to MLLM. 
To bridge this gap, we conduct an in-depth investigation of the attention mechanisms during MLLM inference, using LLaVA as a case study. Our analysis reveals several intriguing phenomena that have not been observed in previous empirical studies~\cite{fastv,VTW}.
For example, Fig. \ref{fig_2}d shows that the attention within the visual token changes dramatically as the layers deepen: in the shallow layers, the attention is distributed uniformly across the visual tokens, whereas in the deep layers, the model selectively discards most visual tokens, focusing the attention on a subset of tokens. 
Additionally, Fig. \ref{fig_2}b shows that as the length of the generated text tokens increases, the attention received by visual tokens gradually diminishes, indicating a decreasing reliance on visual features in the later stages of the generation process. Furthermore, by examining the cosine similarity between the attention of the different layers on various datasets, we find that the attention patterns between adjacent layers are highly similar (Fig.~\ref{fig_cos_qk} and Fig.~\ref{fig_2}c). We refer to this phenomenon as ``lazy layer'', where the subsequent layer essentially mimics the attention pattern of the preceding layer. 

By incorporating the insights gained from our empirical studies, we propose an effective MLLM token pruning framework for accelerating MLLM inference.
We first introduce a Progressive Visual Token Pruning (\textbf{PVTP}) mechanism, which progressively prunes inattentive visual tokens as the model's layers deepen. 
Inspired by the ``lazy layer'' phenomenon, \textbf{PVTP} leverages a \textit{step-wise} mechanism to perform token pruning, ensuring the same visual tokens are reused across the layers between two pruning executions. 
Furthermore, we introduce a method called Visual Token Annealing (\textbf{VTA}) during the generation process, which dynamically adjusts the number of retained visual tokens in the current decoding process based on the length of the generated text sequence.
By combining {\bf PVTP} and {\bf VTA}, our approach differs significantly from previous token pruning techniques, as illustrated in Fig.~\ref{fig_token_prune_compare}.
Our method maintains the quality of generated outputs and requires significantly fewer FLOPs, while achieving a lower decoding latency compared to existing methods on a wide range of multimodal tasks.

In summary, the main contributions of this paper can be outlined as follows:
\begin{itemize}
\item We conduct a comprehensive analysis of the visual attention, systematically elucidating the basis for principled and effective visual token pruning in MLLMs.
\item We propose a novel, retraining-free acceleration framework for MLLMs, which achieves a more than 50\% FLOPs reduction for LLaVA and improves its decoding speed by 2$\times$, without significant impact on accuracy.
\item We validate the superiority of our method over existing methods and quantify the impact of our proposed techniques across a wide range of multimodal task datasets.
\end{itemize}

\section{Related Work}
\textbf{Multimodal Large language models.} The explosive growth of Large Language Models (LLMs)~\cite{gpt4,qwen,vicuna,llama2} has promoted rapid advancements of Multimodal Large Language Models (MLLMs), which combine the language understanding capabilities of LLMs with the visual processing abilities of vision models. 
These MLLMs have demonstrated remarkable success in tasks requiring the integration of textual and visual information, such as visual question answering, image captioning, and multimodal reasoning. 
The pioneering model CLIP~\cite{CLIP} has taken a big stride in bridging the gap between the visual and textual modalities. 
Subsequent works~\cite{flamingo,blip,blip2,llava,llava1.5,llava-next} have built upon the CLIP model, using its visual encoder to extract visual tokens. 
They leverage the wealth of image-text paired datasets and the power of cross-modal alignment and optimization, resulting in a considerable enhancement of learning efficiency. 
This progression signifies an important advancement in the realm of MLLMs, expanding the range of applications by holistically embracing both textual and visual modalities. 
Other impactful works include Gemini~\cite{gemini,gemini1.5}, MiniGPT4~\cite{minigpt}, InternVL~\cite{internvl}, CogVLM~\cite{cogvlm} and InternLM-XComposer~\cite{internlm,internlm2,internlm2.5,internlm4k}. 
While these works have been important drivers in the thriving development of the MLLMs field, our work focuses on exploring the redundant tokens and computations in MLLMs (e.g. LLaVA), with the goal of improving their inference efficiency. \\
\textbf{Vision Token Compression For MLLMs.} 
To reduce the unnecessary computational cost caused by the bloated visual tokens while effectively aligning the visual modality with the text modality, many methods~\cite{flamingo,qwen-vl,blip2,feast,honeybee,tokenpacker} adopt a projector between the visual encoder and the LLM. 
Resampler~\cite{qwen-vl} and Q-former~\cite{blip,blip2} use cross-attention that transfers the visual information to a few learnable tokens. 
Honeybee~\cite{honeybee} and MobileVLM~\cite{mobilevlm,mobilevlm2} leverage convolution layers to aggregate local features and generate the compressed tokens. 
DeCo~\cite{Deco} employs a naive pooling technique to downsample the visual features. 
SeTok~\cite{SeTok} consolidates tokens carrying similar semantics. 
AcFormer~\cite{visual_anchor} employs attention focus of the visual encoder as anchors to drive the information aggregation. 
VoCo~\cite{voco—llama} compresses tokens via intrinsic token distillation. 
LLaVolta~\cite{llavolta} incorporates stage-wise visual context compression to progressively compress the visual tokens from heavily to lightly. 
These methods reduce the number of visual tokens, but they all require additional training of the MLLMs. 
To alleviate this overhead, several training-free methods have emerged. 
CrossGet \cite{crossget} merges visual tokens through Complete-Graph Soft Matching.
LLaVA-Prumerge~\cite{purmaerge} performs adaptive tokens merging before feeding them into the LLM, but ignores the necessity of modality interactions.
FastV~\cite{fastv} and VTW~\cite{VTW} leverage the attention scores within the LLM decoder to evaluate the importance of each token, and then prune the inattentive tokens. 
However, their analysis on attention is not comprehensive, lacking a dynamic, holistic observation of the entire generation process. 
Compared to these methods, we conduct a more systematic analysis of the attention in MLLMs and propose a training-free dynamic pruning technique.

\begin{figure*}[t]
\centering
\includegraphics[width=1\textwidth]{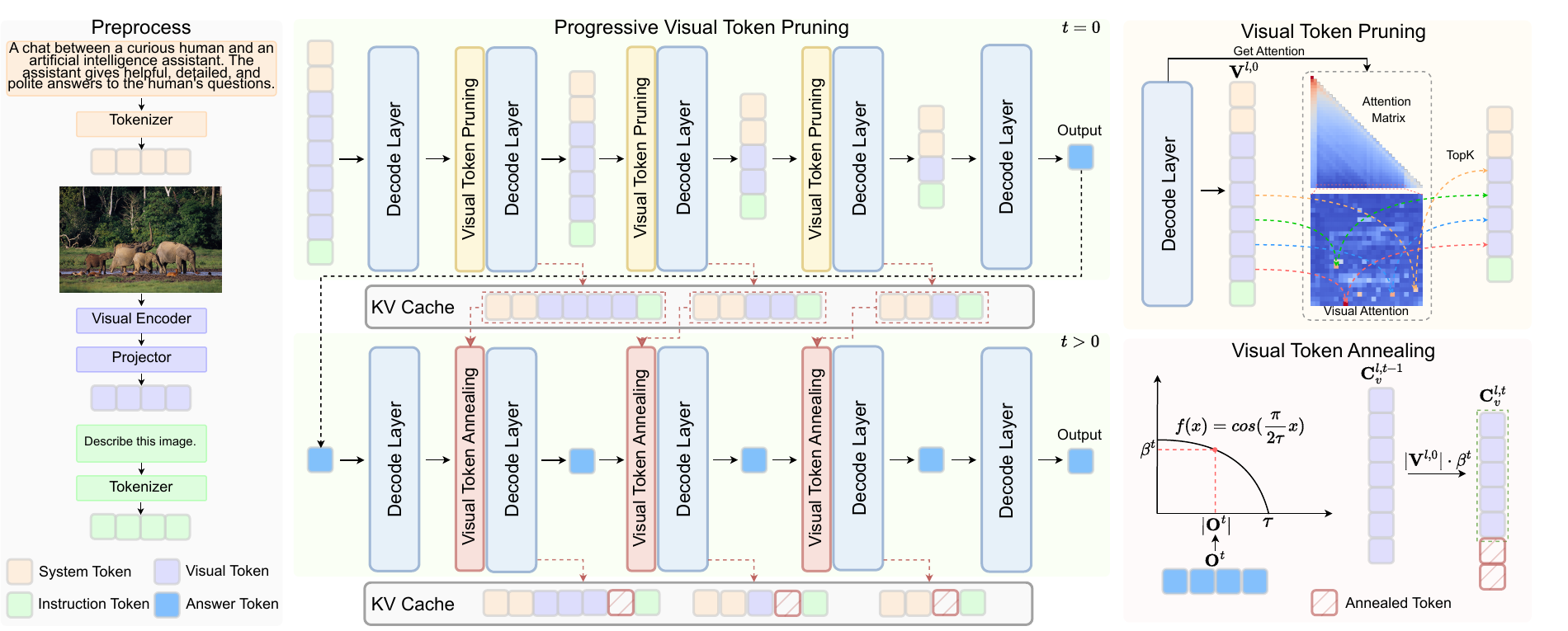}
\caption{Overview of ST$^3$ framework. \textbf{Preprocess} first converts the input from various modalities into tokens. All tokens are concatenated and fed into the LLM's decoder layers. \textbf{Progressive Visual Token Pruning (PVTP)} gradually prunes away non-critical visual tokens throughout the entire decoding forward process. Top-right illustrates the details of \textbf{Visual Token Pruning}. It extracts the visual token attention from the attention matrix of the previous decoding layer, and selectively retains the TopK most important tokens. The predicted token is then used as the input for the next generation step. \textbf{Visual Token Annealing} (\textbf{VTA}) employs a cosine function to control the decay of visual token KV cache \cite{kvcache} in each generation step.}
\label{fig_prune_pipeline}
\end{figure*}

\section{Methodology}
\subsection{Preliminary}

In this work, we use LLaVA~\cite{llava,llava1.5,llava-next}, a widely-adopted model with an exemplar MLLM architecture, to demonstrate our proposed visual token pruning strategy. 
LLaVA is composed of three pivotal components: 1) Visual Encoder, which employs the CLIP-ViT-L/14~\cite{CLIP, ViT} as its backbone to convert an input image into visual embedding; 2) Visual Projector, which maps the visual embedding to the text embedding space through a fully connected layer; 3) LLM, which takes visual tokens and text tokens as input, and predicts the next text token in an autoregressive manner. 
After preprocessing different modalities input, all tokens are concatenated and the input of layers in LLM can then be represented as:
\begin{equation}
\mathbf{X}^{l,t} = [\mathbf{S}, \mathbf{V}^{l,t}, \mathbf{I}, \mathbf{O}^t]
\end{equation}
where $\mathbf{X}^{l,t}$ and $\mathbf{V}^{l,t}$ denote the inputs and
visual tokens of the $l$-th layer in $t$-th generation, $\mathbf{S}$ and $\mathbf{I}$ are system tokens and instruction tokens, and $\mathbf{O}^t$ is the output token at $t$-th generation initialized as $\emptyset$. 
The pipeline of our method is illustrated in Fig. \ref{fig_prune_pipeline}.

\subsection{Progressive Visual Token Pruning}
Token pruning~\cite{tome,algm,ppt} techniques are well-developed for ViTs, and these approaches are often able to effectively retain the most crucial tokens. However, these methods typically incur a significant computational overhead, and their effectiveness on MLLMs remains unclear.
In this paper, we propose an approach based on attention score ranking to screen pivotal tokens, which has been proven effective in FastV~\cite{fastv}. 
Specifically, attention scores are derived from the causal self-attention operation within a decoder layer in LLMs. 
The self-attention in $l$-th layer can be formulated as:
\begin{equation}
    \text{Self-attn}(Q^{l,t},K^{l,t},V^{l,t}) = A^{l,t}V^{l,t}
\label{eq:3}
\end{equation}
where $A^{l,t} = \text{Softmax} \left(\frac{Q^{l,t}(K^{l,t})^T+M}{\sqrt{d}}\right)$, $Q^{l,t}$, $K^{l,t}$ and $V^{l,t}$ are query, key and value projected from $X^{l,t}$, and $M$ is an upper triangular matrix where all non-zero elements are set to $-\infty$ and the diagonal elements are set to 0. 

Considering that the model relies solely on the last token to predict the next output token autoregressively~\cite{VTW}, we take its last row as the attention score $s^{l,t}$ and extract the attention scores on visual tokens $\alpha^{l,t}$. 
By sorting the $\alpha^{l,t}$, we retain visual tokens with higher importance:
\begin{equation}
\mathbf{V}^{l,t} = \mathbf{V}^{l,t}[\text{Top}(G,\alpha^{l-1,t}_i)] \quad G=|\mathbf{V}^{l,t}|-N;t=0
\label{eq:4}
\end{equation}
where $\alpha^{l-1,t}_i$ means attention score of $i$-th visual token, Top($G$,*) denotes indices of highest $G$ values, $|\mathbf{V}^{l,t}|$ is the length of $\mathbf{V}^{l,t}$, and $N$ is the number of pruned visual token.

The strategy we adopt is tailored to the model's empirical behaviors and aligns well with our analysis of the attention mechanism in MLLMs (as in Fig.~\ref{fig_2}d):
In the shallow layers, the attention on visual tokens is quite dense, and the model requires a larger number of visual tokens to gather information;
as the depth increases, the visual attention becomes sparser, which means the model focusing on a smaller set of tokens. 
Therefore, to preserve the crucial shallow-level information, we avoid pruning tokens in the first three layers.

\begin{figure}[t]
\centering
\includegraphics[width=0.47\textwidth]{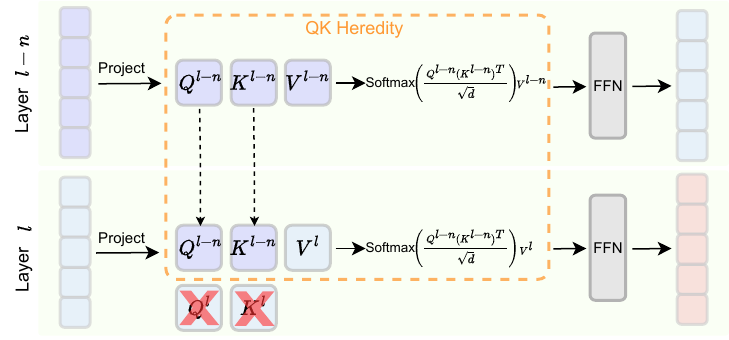}
\caption{QK heredity. Lazy layers $l$ inherits the $Q^{l-n}$ and $K^{l-n}$ from the shallower layer.}
\label{fig_qk_heredity}
\end{figure}

Moreover, we further conduct an in-depth analysis and statistical evaluation of the attention score $s^{l,t}$. 
Specifically, we calculate the cosine similarity $c^{l,t}$ of the attention score distributions between each pair of adjacent layers. This can be expressed by the following formula: 
\begin{equation}
c^{l,l-1,t}= \text{Cosine-similarity}(s^{l,t},s^{l-1,t}) \quad l\in[2,32]
\label{eq:5}
\end{equation}
Based on observation (Fig. \ref{fig_2}c) of $s^{l,t}$, we find the distributions of attention scores between each layer $l$ and its previous layer $l-1$ are similar. 
Building upon the insights gained from this, we hypothesize that a similar pattern may exist across the multi layers as well. 
To validate this, we further expand Eq. \ref{eq:5}:
\begin{equation}
c^{l,l^{'},t}= \text{Cosine-similarity}(s^{l,t},s^{l',t}) \quad l,l^{'}\in[2,32]
\end{equation}
where $c^{l,l^{'},t}$ refers to cosine similarity of the attention score distributions between all layers. 
As shown in Fig.~\ref{fig_cos_qk}, the attention scores between adjacent layers indeed possess a shared pattern.
In other words, we find ``lazy'' layers which are essentially mimicking their preceding layers.
Based on these findings, we introduce two techniques to exploit this ``lazy layer'' phenomenon and reduce the computational overhead: 1) a strided pruning approach that skips over layers instead of pruning every layer, which reduces Top($K$,*) computation, and 2) QK heredity at certain layers which reduces attention computation by inheriting the attention pattern from previous layers.
Specifically, we formulate the attention operation in a ``lazy layer'' $l$ as:
\begin{equation}
\text{Attention}(Q^l,K^l,V^l) = \text{Self-attn}(Q^{l-n},K^{l-n},V^l)
\end{equation}
We let the ``lazy layer'' inherit the $Q^{l-n}$, $K^{l-n}$ and $\text{Softmax}\left(Q^{l-n}(K^{l-n})^T/\sqrt{d}\right)$ results from the previous layer. 
Fig.~\ref{fig_qk_heredity} illustrates the principle of QK heredity. 

\setlength{\tabcolsep}{4pt}
\begin{table*}[t]
\footnotesize
\renewcommand{\arraystretch}{0.985}
\setlength{\tabcolsep}{2mm}
\centering
\begin{tabular}{l|cccccccc}
\noalign{\hrule height 1pt}
Methods & FLOPs$\downarrow$ & Latency$\downarrow$ & MME$\uparrow$ & AI2D$\uparrow$ & SQA$\uparrow$ & MMMU$\uparrow$ & MMB$\uparrow$ & POPE$\uparrow$ \\ 
\hline
LLaVA-1.5-7B & 9.38T & 70.80ms & 1861.97 & 55.25 & 69.46 & 35.20 & \textbf{64.00} & \textbf{86.99} \\
FastV$\ddagger$~\cite{fastv} & 9.38T\decrease{00.00\%} & 70.80ms\decrease{00.00\%} & 1862.12 & 55.34 & 68.77 & 35.00 & 63.83 & 85.13 \\
FastV$\dagger$~\cite{fastv} & 5.78T\decrease{38.38\%} & 42.38ms\decrease{40.14\%} & 1854.87 & 55.38 & 68.77 & 35.10 & 63.83 & 85.13 \\
VTW~\cite{VTW} & 5.38T\decrease{42.64\%} & 46.24ms\decrease{34.69\%} & 1849.51 & \textbf{55.41} & \textbf{69.61} & \textbf{36.10} & 63.92 & 86.89 \\
\rowcolor{violet!10}
ST$^3$ (Ours) & \textbf{4.37T}\decrease{\textbf{53.41\%}} & \textbf{40.12ms}\decrease{\textbf{43.33\%}} & \textbf{1866.71} & \textbf{55.41} & 68.96 & 35.30 & 63.83 & 85.28 \\
\hline
LLaVA-1.5-13B & 17.81T & 128.49ms & 1817.95 & 59.26 & 72.78 & 35.00 & \textbf{68.81} & 87.09 \\
FastV$\ddagger$~\cite{fastv} & 17.81T\decrease{00.00\%} & 128.49ms\decrease{00.00\%} & 1817.55 & 58.87 & 72.98 & 33.90 & 68.30 & 86.17 \\
FastV$\dagger$~\cite{fastv} & 10.73T\decrease{39.75\%} & 73.34ms\decrease{42.92\%} & \textbf{1860.51} & 58.84 & 72.88 & 34.70 & 68.56 & 86.67 \\
VTW~\cite{VTW} & 10.14T\decrease{43.07\%} & 80.74ms\decrease{37.16\%} & 1836.18 & \textbf{59.42} & 72.68 & \textbf{35.10} & \textbf{68.81} & \textbf{87.17} \\
\rowcolor{violet!10}
ST$^3$ (Ours) & \textbf{7.88T}\decrease{\textbf{55.76\%}} & \textbf{63.15ms}\decrease{\textbf{50.85\%}} & 1830.51 & 58.81 & \textbf{73.18} & 34.60 & 68.21 & 86.81  \\
\hline
LLaVA-NeXT-7B & 30.95T & 408.61ms & 1850.56 & 65.25 & \textbf{70.15} & 35.40 &\textbf{67.10} & 87.62 \\
FastV$\ddagger$~\cite{fastv} & 30.95T\decrease{00.00\%} & 408.61ms\decrease{00.00\%} & 1806.93 & 65.12 & 69.21 & 35.70 & 66.41 & \textbf{87.78}\\
FastV$\dagger$~\cite{fastv} & 17.83T\decrease{42.39\%} & 207.34ms\decrease{49.26\%} & 1804.95 & 64.64 & 69.21 & 35.10 & 66.32 & 87.47\\
VTW~\cite{VTW} & 16.74T\decrease{45.91\%} & 229.81ms\decrease{43.76\%} & \textbf{1852.56} & \textbf{65.41} & 70.00 & \textbf{35.80} & \textbf{67.10} & 87.60 \\
\rowcolor{violet!10}
ST$^3$ (Ours) & \textbf{13.87T}\decrease{\textbf{55.19\%}} & \textbf{148.85ms}\decrease{\textbf{63.57\%}} & 1822.42 & 64.51 & 69.06 & 35.10 & 66.67 & 87.39  \\
\hline
LLaVA-NeXT-13B & 57.66T & 691.02ms & \textbf{1902.39} & 70.27 & 73.57 & \textbf{36.10} & \textbf{69.33} & \textbf{87.53} \\
FastV$\ddagger$~\cite{fastv} & 57.66T\decrease{00.00\%} & 691.02ms\decrease{00.00\%} & 1887.00 & 69.95 & 72.98 & 35.90 & 68.81 & 87.48 \\
FastV$\dagger$~\cite{fastv} & 32.19T\decrease{44.17\%} & 346.71ms\decrease{49.83\%} & 1874.82 & 69.98 & 73.03 & 35.90 & 68.47 & 87.30 \\
VTW~\cite{VTW} & 30.61T\decrease{46.91\%} & 363.98ms\decrease{47.33\%} & 1879.96 & 70.24 & 73.43 & 35.30 & 69.07 & 87.44 \\
\rowcolor{violet!10}
ST$^3$ (Ours) & \textbf{22.02T}\decrease{\textbf{61.81\%}} & \textbf{218.71ms}\decrease{\textbf{68.35\%}} & 1886.23 & \textbf{70.29} & \textbf{73.63} & 35.30 & 69.07 & 87.44 \\
\noalign{\hrule height 1pt}
\end{tabular}
\caption{Comparison of existing training-free token pruning method on MLLMs with single-token answer datasets. SQA means the ScienceQA\cite{SQA} image subset, MMMU represents the validation subset of MMMU\cite{mmmu}, and MMB denotes the english subset of MMBench\cite{mmbench}. Latency refers to the total time required for the forward propagation to pass through all decode layers. ``$\ddagger$'' means attention mask implementation and ``$\dagger$'' means token drop implementation in its code base. The best results are in \textbf{bold}.}
\label{table:1}
\end{table*}

\subsection{Visual Token Annealing}
Based on another finding from our experiments: the proportion of visual attention decreases as the generated text tokens increase (as in Fig.~\ref{fig_2}b), we propose a strategy called Visual Token Annealing (VTA). VTA dynamically adjusts the quantity of KV cache corresponding to visual tokens retained at each layer based on generated text tokens over the whole autoregressive process. VTA can be formulated as:

\begin{equation}
\mathbf{C}_v^{l,t} = \mathbf{C}_v^{l,t-1}\Bigl[:|\mathbf{V}^{l,0}|\cdot\beta^t\Bigl] \quad t>0 
\end{equation}

\begin{equation}
\beta^{t} =\left\{
\begin{array}{lcl}
\text{cos}(|\mathbf{O}^t|\cdot\frac{\pi}{2\tau})        &        & |\mathbf{O}^t|<\tau\\
0       &        &|\mathbf{O}^t|\geq\tau  \\
\end{array} \right.
\end{equation}
where $\mathbf{C}_v$ means KV cache of visual tokens, $|\mathbf{O}^t|$ denotes length of $\mathbf{O}^t$, and $\tau$ is a hyper-parameter controlling the rate of decay in visual tokens while enforcing a truncation on the maximum generation length.
The rationale for selecting the cosine function for VTA is based on the following two considerations when $|\mathbf{O}^t|\in[0,\tau]$: 1) the first derivative of the cosine function is negative, ensuring gradual shrinkage of visual tokens as the text tokens are generated; 2) its second derivative is also negative, reinforcing a larger compression of visual tokens in the later stages of the generation process.

\begin{table}[t]
\centering
\footnotesize
\renewcommand{\arraystretch}{1}
\setlength{\tabcolsep}{0.5mm}
\begin{tabular}{lccccccc}
\noalign{\hrule height 1pt}
Methods & Memory$\downarrow$ &Coco2017$\uparrow$ & Flickr30k$\uparrow$ & Nocaps$\uparrow$ \\ 
\hline
LLaVA-1.5-7B & 4.94G & 110.43 & 74.89 & 105.53 \\
FastV$\ddagger$ & 4.94G & \textbf{110.80} & 74.70 & 105.36\\
VTW & 2.47G & 67.20 & 40.65 & 95.78\\
\rowcolor{violet!10}
ST$^3$ (Ours) & \textbf{2.27G} & 110.66 & \textbf{74.91} & \textbf{105.62}\\
\hline
LLaVA-1.5-13B & 7.70G & 115.57 & 79.56 & \textbf{109.31}\\
FastV$\ddagger$ & 7.70G & 115.91 & 79.68 & 108.85\\
VTW & 3.75G & 101.67 & 65.87 & 95.21\\
\rowcolor{violet!10}
ST$^3$ (Ours) & \textbf{2.51G} & \textbf{116.33} & \textbf{79.86} & 108.45\\
\hline
LLaVA-NeXT-7B & 15.68G & \textbf{99.87} & \textbf{68.47} & \textbf{88.37}\\
FastV$\ddagger$ & 15.68G & 98.54 & 67.18 & 86.48 \\
VTW & 7.84G & 82.50 & 57.81 & 43.76 \\
\rowcolor{violet!10}
ST$^3$ (Ours) & \textbf{7.21G} &  98.57 & 66.93 & 86.29\\
\hline
LLaVA-NeXT-13B & 24.48G & \textbf{101.93} & \textbf{66.70} & \textbf{88.18}\\
FastV$\ddagger$ & 24.48G & 101.29 & 66.13 & 87.86 \\
VTW & 12.24G & 86.86 & 56.52 & 43.26\\
\rowcolor{violet!10}
ST$^3$ (Ours) & \textbf{7.98G} & 101.37 & 65.99 & 87.96\\
\noalign{\hrule height 1pt}
\end{tabular}
\caption{Comparison of existing training-free token pruning method on MLLMs with image caption datasets. ``Memory'' means KV cache memory, which is calculated over the same sample after first token is generated (batchsize=16). ``Coco2017'' refers to validation subset of COCO2017 caption\cite{coco}. ``Flickr30k'' and ``Nocaps'' are validation and test splits in the original datasets\cite{flickr30k,nocaps}. The best results are in \textbf{bold}.}
\label{table:2}
\end{table}

\begin{table}[t]
\centering
\footnotesize
\renewcommand{\arraystretch}{1}
\setlength{\tabcolsep}{2.7mm}
\begin{tabular}{lcccccc}
\noalign{\hrule height 1pt}
\multicolumn{6}{c}{LLaVA-1.5-7B ($C=1\%$)} \\
\hline
$S$ & $R$ & FLOPs & Latency & MME & AI2D  \\ 
\hline
1& 1.75\% & \textbf{4.03T} & 43.88ms & 1836.47 & 54.63 \\
2& 3.50\% & 4.08T & 41.05ms & 1862.15 & 54.53\\
4& 7.00\% & 4.20T & 40.74ms & 1840.29 & 54.99\\
\rowcolor{violet!10}
7& 12.25\% & 4.37T & \textbf{40.12ms} & \textbf{1866.71} & \textbf{55.41} \\
14& 24.50\% & 4.78T & 40.24ms & 1861.87 & 55.41 \\
28& 49.00\% & 5.59T & 42.67ms & 1858.10 & 55.31 \\
\noalign{\hrule height 1pt}
\multicolumn{6}{c}{LLaVA-1.5-13B ($C=5\%$)} \\
\hline
$S$ & $R$ & FLOPs & Latency & MME & AI2D  \\ 
\hline
1 & 1.25\% & \textbf{7.64T} & 68.28ms & 1826.94 & 58.78 \\
2 & 2.50\% & 7.72T & 63.96ms & \textbf{1851.51} & 58.91 \\
3 & 3.75\% & 7.80T & 64.54ms & 1833.87 & 59.10 \\
\rowcolor{violet!10}
4 & 5.00\% & 7.88T & \textbf{63.15ms} & 1830.51 & 58.81 \\
6 & 7.50\% & 8.04T & 63.34ms & 1827.58 & \textbf{59.07} \\
9 & 11.25\% & 8.29T & 64.06ms & 1832.09 & 58.71 \\
\noalign{\hrule height 1pt}
\end{tabular}
\caption{Results of the pruning stride ``$S$'' and the pruning ratio ``$R$'' per step. The best results are in \textbf{bold}.}
\label{table:3}
\end{table}

\section{Experiments}

\subsection{Comparison on Single-token Answer Datasets}
We compare our method with existing methods on mainstream single-token answer datasets, with the results shown in Tab. \ref{table:1}. Among them, for the LLaVA-NeXT model, we use the LLaVA-1.6-vicuna version. 
Compared to the original LLaVA series models, our method reduces more than 50\% FLOPs and achieves over 2-3$\times$ improvements in inference speed, with almost no loss in accuracy. 
Additionally, compared to the current state-of-the-art methods, our method shows clear advantages in terms of inference speed and computational cost, while maintaining comparable or even better accuracy. 
The experimental results are consistent with our observations that MLLMs tend to focus their attention on a few tokens at deeper layers. 
Thus, as layers deepen, only fewer tokens need to be retained. 
This property allows us to drastically improve inference efficiency by pruning a large ratio of tokens without compromising accuracy.

\subsection{Ablation on Progressive Visual Token Pruning}
We conduct ablation experiments on pruning parameters stride $S$ and pruning ratio $R$. 
The proportion of conserved visual tokens in the last layer $C$ can be calculated as:
\begin{equation}
C=1-\lfloor(L-3)/S\rfloor\cdot R-P
\end{equation}
where $P$ means pruning ratio in $4$-th layer, which is set to $50\%$ inspired by FastV~\cite{fastv}, and ``$\lfloor*\rfloor$'' is the round-down operation. 
Tab. \ref{table:3} shows the results when $C=1\%$ on LLaVA-1.5-7B and $C=5\%$ on LLaVA-1.5-13B. 
Lower FLOPs cannot achieve the lowest latency as a smaller pruning stride incurs a larger overhead. 
Thus, in practice, the selection of $S$ and $R$ is optimized with the primary objective of attaining the minimal latency with competitive accuracy. 
Incorporating QK heredity (Fig.~\ref{fig_qk_heredity}) along with PVTP can further improve acceleration performance.
Tab.~\ref{table:6} shows the effect of implementing QK heredity at different layers: in shallow layers, it leads to a decrease in accuracy, while the performance is robust when it is applied to later layers. 
We believe this is due to the errors introduced by QK heredity compounding as layers deepen.

\begin{figure}[t]
\centering
\includegraphics[width=0.47\textwidth]{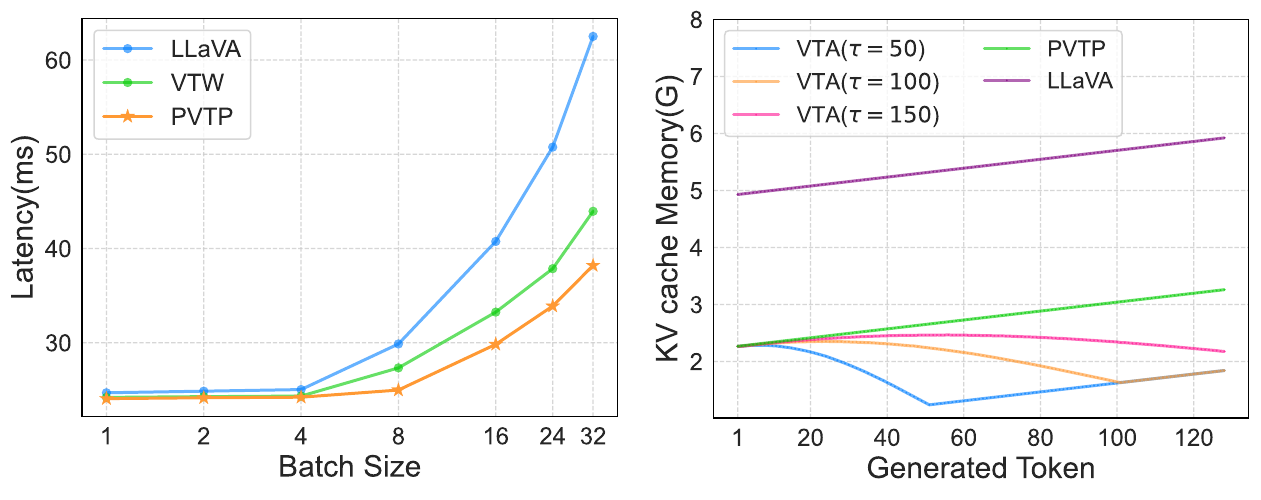}
\caption{\textit{Left}: Changes in decode latency under distinct batch sizes. \textit{Right}: Relationship between KV cache Memory, decode latency and the length of generated sequence, with imagesize=336*336 and batchsize=16.}
\label{fig_latency}
\end{figure}

\begin{table}[t]
\centering
\scriptsize
\footnotesize
\renewcommand{\arraystretch}{1}
\setlength{\tabcolsep}{0.5mm}
\begin{tabular}{lccccccc}
\noalign{\hrule height 1pt}
Methods & FLOPs$\downarrow$ & Latency$\downarrow$ & MME$\uparrow$ & AI2D$\uparrow$ & SQA$\uparrow$ & MMMU$\uparrow$ \\ 
\hline
ST$^3$ & 4.37T & 40.12ms & 1866.71 & 55.41 & 68.96 & 35.30 \\
7$\rightarrow$8$\sim$10 & 4.30T & 38.66ms & 1480.70 & 43.10 & 62.47 & 34.20 \\
14$\rightarrow$15$\sim$17 & 4.31T & 38.89ms & 1825.35 & 49.32 & 62.62 & 32.10 \\
21$\rightarrow$22$\sim$24 & 4.33T & 39.15ms & 1864.18 & 54.70 & 69.01 & 36.30 \\
\rowcolor{violet!10}
28$\rightarrow$29$\sim$31 & 4.34T & 38.88ms & 1863.54 & 55.25 & 69.06 & 35.70 \\
\noalign{\hrule height 1pt}
\end{tabular}
\caption{Results of implementing QK heredity (Fig.~\ref{fig_qk_heredity}) across multiple layers. ``4$\rightarrow$5$\sim$7'' means reusing attention score of layer4 from layer5 to layer7 based on ST$^3$.}
\label{table:6}
\end{table}

\subsection{Comparison on Long-sentences Answer Datasets}
For longer sentence generation, we employ the widely adopted  KV cache~\cite{kvcache} technique. 
After the first token generation, subsequent token generations directly read the Key-Value from the KV cache for the attention computation. 
Therefore, the implementation of Visual Token Annealing (VTA) actually acts on the KV cache by evicting the Key-Value pairs corresponding to the visual token positions. 
Based on extensive experimentation, we find: (1) It is unnecessary to \textit{retain the entire KV cache of visual tokens in the first generation}, and (2) KV cache of Visual tokens can be \textit{gradually discarded as the generation unfolds}. 
Tab. \ref{table:2} demonstrates that our method can achieve a comparable generation quality, while only requiring less than 50\% or even 30\% of the KV cache memory compared to LLaVA. 
Notably, VTW~\cite{VTW} fails on long-token generation datasets while it performs well on single-token datasets, showing our method's robustness in comparison.
In the left of Fig. \ref{fig_latency}, we show the results from testing the latency of generating one token for different methods with KV cache under varying batchsizes. 
PVTP exhibits lower latency compared to existing methods across various settings. The right of Fig. \ref{fig_latency} illustrates the relationship between the length of the generated sequence and the KV cache memory usage. 
It shows that VTA maintains a low memory usage throughout the entire generation pipeline. The kinks in the blue and orange line represent the points where all the cache corresponding to visual tokens is cleared at that node. 

\begin{figure}[H]
\centering
\includegraphics[width=0.47\textwidth]{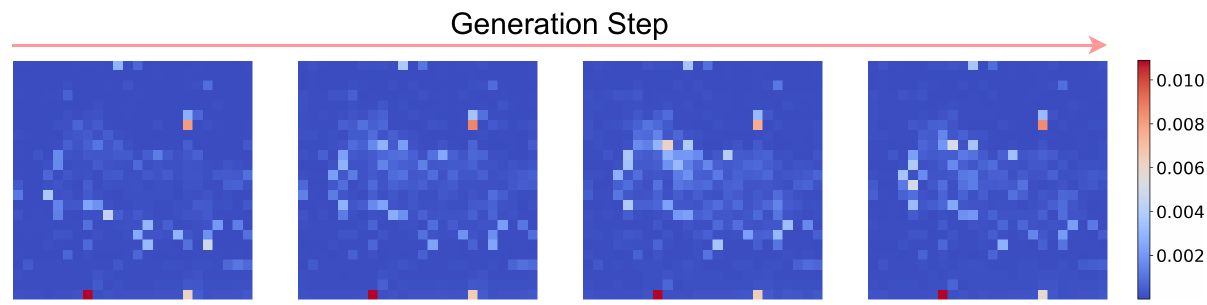}
\caption{Attention on visual tokens in different generation step. The distribution of visual attention is similar across the generation process.}
\label{fig_attn_step}
\end{figure}

To understand why partial KV cache is enough, we note that: 1) as shown in Tab. \ref{table:overlap}, a large proportion of the Top 50\% attention visual tokens overlaps across different steps, suggesting that the KV cache trimmed after the first generation step can be reused in the subsequent generation steps; 2) the distribution of visual token attention remains similar at each generation step (see Fig. \ref{fig_attn_step}). The overlap is defined by:

\begin{equation}
\text{Overlap}_n = \frac{|\textbf{TopK}(T_1)\cap\textbf{TopK}(T_n)|}{|\textbf{TopK}(T_1)|}
\vspace{0.2cm}
\end{equation}
where $T_n$ means the visual tokens in $n$-th generation step. $|\cdot|$ denotes obtaining the number of elements in a set.

\subsection{Ablation on Visual Token Annealing}
Tab. \ref{table:5} provides a comparative analysis of different decay schemes.
The Exponential attenuation function is formulated as $\beta^{t} = e^{-|\mathbf{O}^t|/\sigma}$, and the Linear attenuation function is $\beta^{t} = 1-\frac{|\mathbf{O}^t|}{\tau}$ when $|\mathbf{O}^t|<\tau$ and $0$ otherwise.
The Cosine function demonstrates superior performance compared to the other two, which aligns well with our intuition because it has a lower annealing rate in the initial generation process and thus better preserves the integrity of visual information. 
In contrast, the Exponential function has a faster decay rate in the initial stages and the decay rate decreases uniformly over time, conflicting with the model's reliance on visual tokens in the early generation process. The results reveal that convex functions may be more suitable for VTA. 
A study of the parameter $\tau$ is conducted in the cosine function.
Notably, the accuracy plunges when $\tau=10$, which can be attributed to the mismatch between the $\tau$ and the actual length of the generated tokens: visual tokens are completely discarded once the generated tokens length exceeds 10, resulting in the loss of image information in the subsequent generation process.
Therefore, we recommend setting $\tau$ to be greater than the preset maximum generation length.

\setlength{\tabcolsep}{4pt}
\begin{table}[H]
\centering
\footnotesize
\renewcommand{\arraystretch}{1}
\setlength{\tabcolsep}{1.8mm}
\begin{tabular}{l|c|c|c|c|c}
\noalign{\hrule height 1pt}
Step & 1 & 2 & 3 & 4 & 5 \\
Overlap & 100.00\% & 90.97\% & 82.29\% & 84.72\% & 86.46\% \\
\hline
Step & 6 & 7 & 8 & 9 & 10\\
Overlap & 88.89\% & 82.99\% & 86.81\% & 88.19\% & 88.54\% \\
\noalign{\hrule height 1pt}
\end{tabular}
\caption{Ratio of Top 50\% attention visual token in each generation step that overlap with the first step.}
\label{table:overlap}
\end{table}

\begin{table}[H]
\centering
\footnotesize
\renewcommand{\arraystretch}{1}
\setlength{\tabcolsep}{3mm}
\begin{tabular}{lcccccc}
\noalign{\hrule height 1pt}
Function  & Coco2017 & Flickr30k & Nocaps  \\ 
\hline
Exp. ($\sigma=20$)& 110.29 & 73.65 & 103.55 \\
Linear ($\tau=50$)& 110.50 & 74.33 & 104.59 \\
\rowcolor{violet!10}
Cosine ($\tau=50$)& \textbf{110.66} & \textbf{74.91} & \textbf{105.62} \\
Cosine ($\tau=100$) & 110.58 & 74.47 & 104.96 \\
Cosine ($\tau=75$) & 110.55 & 74.49 &  104.99\\
Cosine ($\tau=25$) & 110.49 & 74.27 & 104.64 \\
Cosine ($\tau=10$) & 103.05 & 64.45 & 96.68 \\ 
\noalign{\hrule height 1pt}
\end{tabular}
\caption{Comparison of the attenuation function with varied parameters on LLaVA-1.5-7B. ``Exp.'' means exponential. The best results are in \textbf{bold}.}
\label{table:5}
\end{table}

\section{Conclusion}
This paper introduces Spatial-Temporal Visual Token Trimming (ST$^3$), a novel framework for pruning visual tokens of Multimodal Large Language Models (MLLM) across both \textit{layer} and \textit{time} dimensions.  
ST$^3$ is based on two key observations from our analysis of MLLM attention mechanisms: numerous visual tokens and partial attention computations are redundant during decoding, and subsequent layers often mimic the attention patterns of preceding ones (``lazy layers"). Experimental results show that ST$^3$ delivers around $\mathbf{2\times}$ faster inference with only about 30\% KV cache memory compared to the
original LLaVA, while maintaining consistent performance across various datasets. 
We anticipate that this work provides valuable empirical insights and advances the field of MLLM token pruning.


\section{Appendix}
\section{Datasets and Implementation}
We validate the proposed strategy across a diverse range of multi-modal tasks, including VQA, visual reasoning and hallucination evaluation. 
Specifically, these datasets are used for evaluating the proposed Progressive Visual Token Pruning (PVTP), because their correct answers are limited to a single token. 
On the other hand, for the evaluation of the Visual Token Annealing (VTA) technique, we adopt the image caption task, in which the answers typically consist of longer sentences. 
In the experiments conducted, the method is applied to popular open-source MLLM architectures, such as LLaVA-1.5-7B and LLaVA-NEXT-7B. 
All experiments are performed on NVIDIA A100 GPU. Details of datasets are shown in Tab. \ref{table:datasets}
\setlength{\tabcolsep}{1.5pt}
\begin{table}[ht]
\centering
\begin{tabular}{l|c|c}
\toprule
Dataset & Task & Metric \\
\midrule
MME & Visual Reasoning & Total Score \\
AI2D & Visual Question Answering & Accuracy \\
ScienceQA\_img & Visual Question Answering & Accuracy \\ 
MMMU\_val & Visual Reasoning & Accuracy \\
MMbench\_en & Visual Reasoning & Accuracy \\
POPE &  Hallucination Evaluation & Accuracy \\
\midrule
Coco2017\_cap\_val & Image Caption & CIDEr \\
Flickr30k\_test & Image Caption & CIDEr \\
Nocaps\_val & Image Caption & CIDEr \\
\bottomrule
\end{tabular}
\caption{Datasets used in the experiments with their tasks and metrics.}
\label{table:datasets}
\end{table}

\section{Details of PVTP}
We conduct extensive ablation experiments on pruning stride and pruning ratio on LLaVA-1.5-7B and LLaVA-1.5-13B. The resualts are shown in Tab. \ref{table:a1} and \ref{table:a2}. We chose $C=1\%$ and $C=5\%$ as the final configurations, because under these parameter settings, the model achieves the optimal trade-off between speed and accuracy. This result further suggests that the model does not need substantial visual tokens any more in deep layers. Alg. \ref{alg:algorithm1} shows the pseudo code of PVTP. It is a simple, effective and plug-and-play method, which can be easily integrated with existing methods with just a few lines of code.

\setlength{\tabcolsep}{8pt}
\begin{table*}[ht]
\centering
\begin{tabular}{lccc|ccc|ccc}
\toprule
\multirow{2}{*}{Datasets}  & \multicolumn{3}{c}{Cosine} & \multicolumn{3}{c}{Linear} & \multicolumn{3}{c}{Exponential}  \\ 
& $\tau$=75 & $\tau$=50 & $\tau$=25 & $\tau$=75 & $\tau$=50 & $\tau$=25 & $\sigma$=20 & $\sigma$=10 & $\sigma$=5 \\
\midrule
Coco2017\_cap\_val& 110.55 & 110.66 & 110.49 & 110.40 & 110.50 & 110.18 & 110.29 & 106.29 & 98.05 \\
Flickr30k\_test& 74.49 & 74.91 & 74.27 & 74.27 & 74.33 & 73.53 & 73.65 & 70.20 & 61.31 \\
Nocaps\_val& 104.99 & 105.62 & 104.64 & 104.78 & 104.59 & 103.51 & 103.55 & 101.26 & 94.20 \\
\bottomrule
\end{tabular}
\caption{Comparison of various attenuation function under different parameters on LLaVA-1.5-7B.}
\label{table:vta_function}
\end{table*}

\section{Details of VTA}
\begin{figure*}[t]
\centering
\includegraphics[width=1\textwidth]{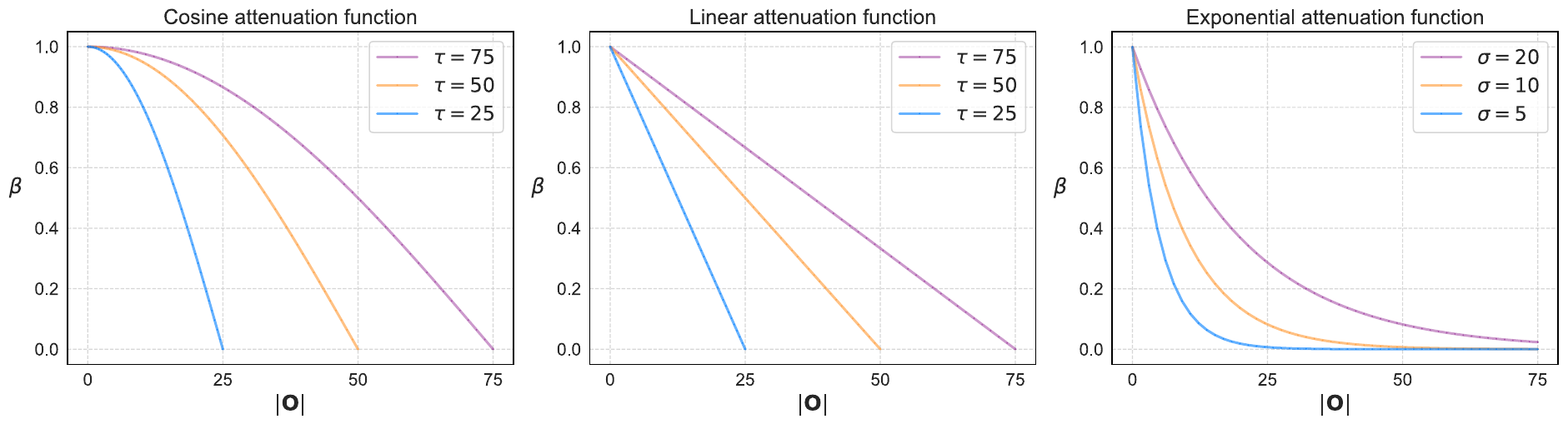}
\caption{Attenuation functions under different parameters.}
\label{fig_vta_function}
\end{figure*}

Alg. \ref{alg:algorithm2} illustrates the pseudo code of VTA under KV cache implementation, which adds a few lines of code based on PVTP framework. We test the generation quality of three attenuation  functions under different parameters. Fig. \ref{fig_vta_function} shows relationship between decay rate and the length of generated sequence. Tab. \ref{table:vta_function} illustrates the superiority of the cosine function, further revealing that convex functions may be more suitable for VTA. 
We conduct a more in-depth quantitative study of the parameter $\tau$ in cosine function, and the results are shown in Tab. \ref{table:4}. 
The model achieves the best performance across three datasets when $\tau=50$. 
Notably, the accuracy plunges when $\tau=10$, which can be attributed to the mismatch between the $\tau$ and the actual length of the generated tokens: the visual tokens are completely discarded once the generated tokens length exceeds 10, resulting in the loss of image information in the subsequent generation process.
Therefore, we recommend setting $\tau$ to be greater than the preset maximum generation length.

\setlength{\tabcolsep}{5.5pt}
\begin{table}[H]
\centering
\begin{tabular}{lcccccc}
\toprule
$\tau$  & Coco2017\_cap\_val & Flickr30k\_test & Nocaps\_val  \\ 
\midrule
100& 110.58 & 74.47 & 104.96 \\
75& 110.55 & 74.49 &  104.99\\
\rowcolor{violet!10}
50& 110.66 & 74.91 & 105.62 \\
25 & 110.49 & 74.27 & 104.64 \\
10 & 103.05 & 64.45 & 96.68 \\
\bottomrule
\end{tabular}
\caption{Ablation for $\tau$ in the annealing function on LLaVA-1.5-7B. The best results are in \textbf{bold}.}
\label{table:4}
\end{table}

\section{More results on QK heredity}
\begin{figure*}[t]
\centering
\includegraphics[width=1\textwidth]{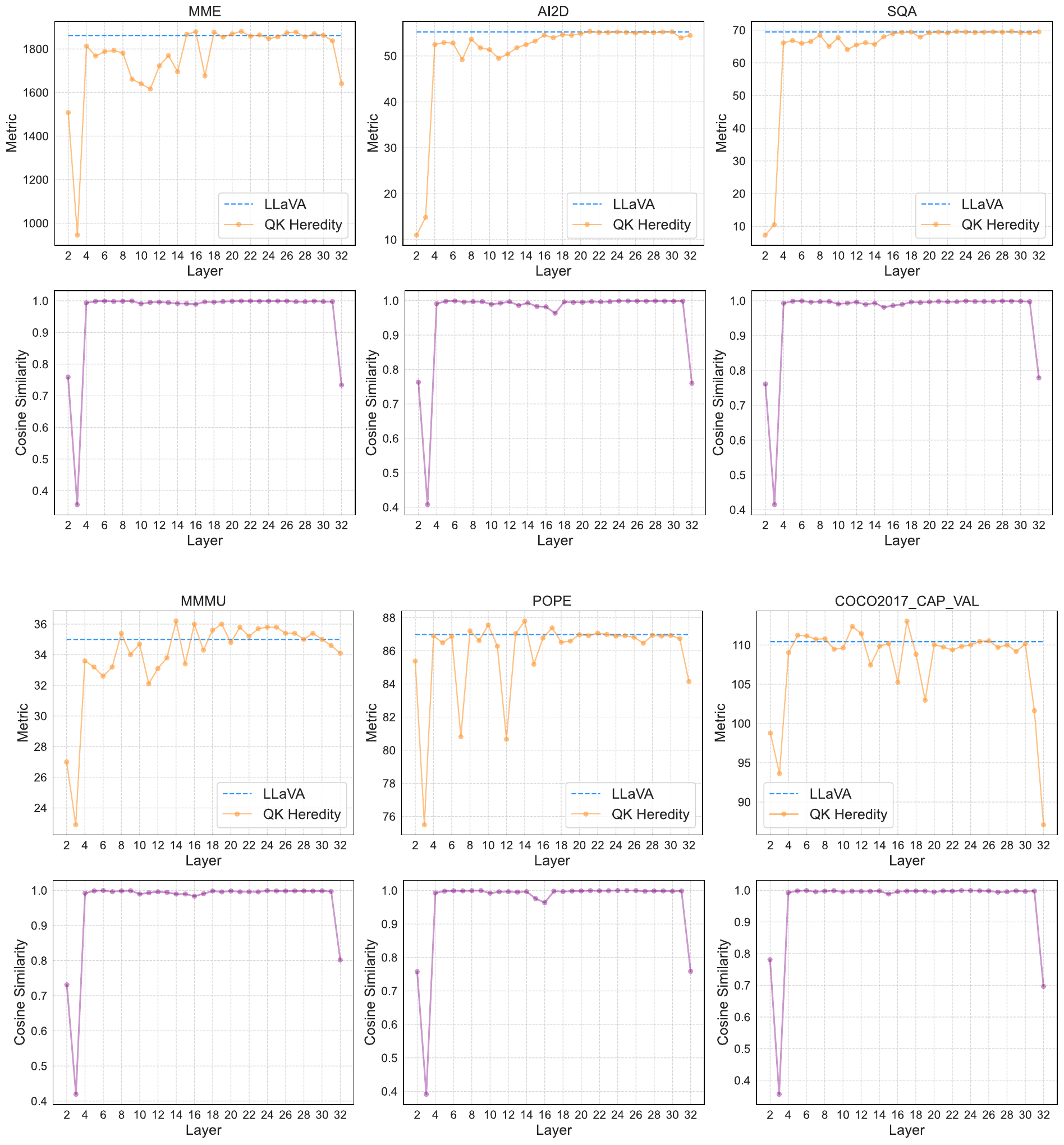}
\caption{\textbf{Top:} Results of implementing QK-heredity ($n=1$) on each layer in LLaVA-1.5-7B over various datasets. \textbf{Bottom:} Cosine similarity of the attention scores between each layer and its previous layer.}
\label{fig_qk_heredity_dataset}
\end{figure*}

\begin{figure*}[t]
\centering
\includegraphics[width=1\textwidth]{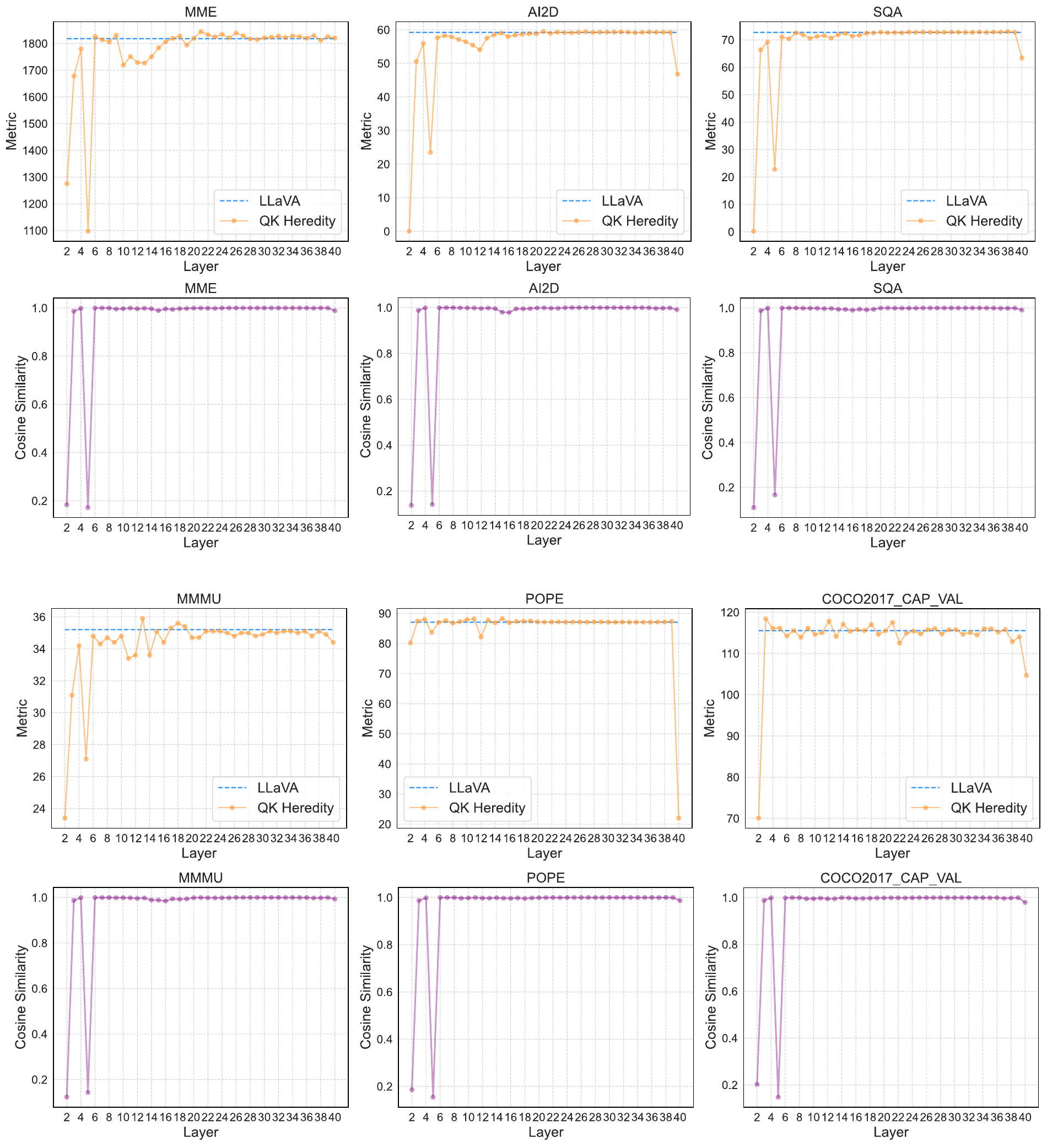}
\caption{\textbf{Top:} Results of implementing QK-heredity ($n=1$) on each layer in LLaVA-1.5-13B over various datasets. \textbf{Bottom:} Cosine similarity of the attention scores between each layer and its previous layer.}
\label{fig_qk_heredity_dataset_13b}
\end{figure*}

We observe the ``lazy layer'' phenomenon in both LLaVA-1.5-7B and LLaVA-1.5-13B as shown in bottom of Fig. \ref{fig_qk_heredity_dataset} and \ref{fig_qk_heredity_dataset_13b}, thus we try to directly apply QK-heredity to the original LLaVA model. Top of Fig. \ref{fig_qk_heredity_dataset} and \ref{fig_qk_heredity_dataset_13b} show the comparison on metric of QK heredity ($n=1$) and LLaVA, indicating that the majority of layers inheriting the attention scores from their previous layers has little impact on the final output (quantitative results can be found in Tab. \ref{table:qk_heredity_datasets7b}and \ref{table:qk_heredity_datasets13b}). In Tab. \ref{table:qk_dataset_multi}, we have conduct a more in-depth exploration of the parameter $n$. The results demonstrate that increasing the parameter $n$ in the deeper layers can provide the model with lossless acceleration. 
\section{Dialogue test}
\begin{figure*}[t]
\centering
\includegraphics[width=1\textwidth]{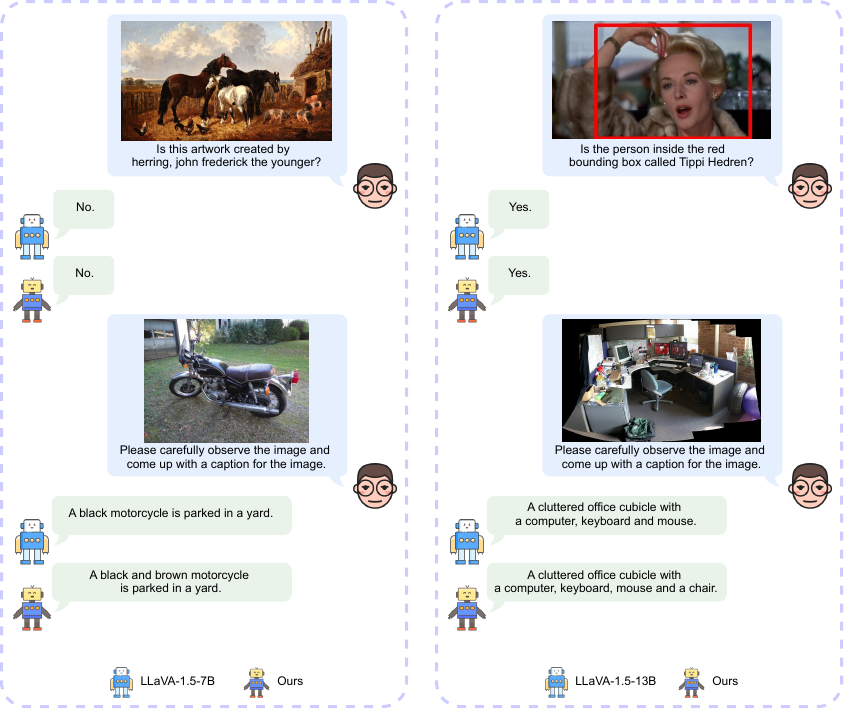}
\caption{Dialogue test on our method and original LLaVA model. Our method exhibits competitive generation quality.}
\label{fig_chat}
\end{figure*}

Please refer to Fig. \ref{fig_chat}.
\section{Limitation}
\begin{itemize}
\item In our PVTP approach, we employed a uniform pruning step size. In the future, we will investigate whether non-uniform step sizes can enhance accuracy.
\item The length of the generated text in the dataset for the experiments is limited. In the future, we will extend the method to datasets with longer texts generation.
\item The current baseline models for this paper are LLaVA 7B and 13B. In the future, we will extend our method to larger LLaVA models and other multimodal large language models.
\end{itemize}

\begin{algorithm}[ht]
\caption{PVTP}
\label{alg:algorithm1}
\textbf{Input}: Hidden\_state $\mathbf{X} = [\mathbf{S}, \mathbf{V}, \mathbf{I}]$\\
\textbf{Parameter}: Pruning stride $S$ and pruning rate $R$; number of layers $L$; visual token attention score $\alpha_v$
\begin{algorithmic}[1] 
\STATE length $=$  $\mathbf{X}.$shape[1]
\FOR{$l$ in Range($L$)}
\IF {$l\geq3$ and $(l-3)\%S=0$ and length $>1$}
\STATE $\mathbf{V} = \mathbf{V}[\text{Topk}((1-R)\cdot|\mathbf{V}|,\alpha_v)]$
\STATE $\mathbf{X} = [\mathbf{S}, \mathbf{V}, \mathbf{I}]$
\ENDIF
\STATE $\mathbf{X}=$ LLaMADecodeLayer($\mathbf{X}$, $l$)
\ENDFOR
\end{algorithmic}
\end{algorithm}

\begin{algorithm}[ht]
\caption{QK heredity in self-attention}
\label{alg:algorithm3}
\textbf{Input}: Hidden\_state $\mathbf{X}$\\
\textbf{Parameter}: Layer index $l$; attention cache $U$ ; lazy layer $W$
\begin{algorithmic}[1] 
\IF {$l \notin W$}
\STATE $Q = \text{q\_proj}(\mathbf{X})$
\STATE $K = \text{k\_proj}(\mathbf{X})$
\STATE $V = \text{v\_proj}(\mathbf{X})$
\STATE $A = QK^{T}$ 
\STATE $U\text{.append}(A)$
\STATE $O = AV$
\ELSE 
\STATE $V = \text{v\_proj}(\mathbf{X})$
\STATE $A = U$[-1]
\STATE $O = AV$
\ENDIF
\end{algorithmic}
\end{algorithm}

\begin{algorithm}[ht]
\caption{PVTP+VTA (KV cache implementation)}
\label{alg:algorithm2}
\textbf{Input}: Hidden\_state $\mathbf{X} = [\mathbf{S}, \mathbf{V}, \mathbf{I}]$ if number of generated tokens is $0$ else $\mathbf{X} = [\mathbf{O}]$\\
\textbf{Parameter}: Number of generated tokens $O$; pruning stride $S$ and pruning rate $R$; number of layers $L$; visual token attention score $\alpha_v$; viusal tokens KV cache $C$; attenuation function F($x$); top index cache $T$
\begin{algorithmic}[1] 
\STATE length $=$  $\mathbf{X}.$shape[1]
\STATE \textcolor{gray}{\#PVTP}
\IF {length $>1$}
\FOR{$l$ in Range($L$)}
\IF {$l\geq3$ and $(l-3)\%S=0$}
\STATE $T$$[~l~]$ $=\text{Topk}((1-R)\cdot|\mathbf{V}|,\alpha_v)$
\STATE $\mathbf{V} = \mathbf{V}[T]$
\STATE $\mathbf{X} = [\mathbf{S}, \mathbf{V}, \mathbf{I}]$
\ENDIF
\STATE $\mathbf{X}=$ LLaMADecodeLayer($\mathbf{X}$,$l$)
\ENDFOR
\ENDIF
\STATE \textcolor{gray}{\#VTA}
\IF{length $=1$}
\FOR{$l$ in Range($L$)}
\IF {$l\geq3$ and $(l-3)\%S=0$}
\STATE $D=$ Cos($O\cdot\frac{\pi}{2\tau}$)
\STATE $T^{'}=T[~l~][:|C[~l~]|\cdot D]$
\STATE $C = C[T^{'}]$
\ENDIF
\STATE $\mathbf{X}=$ LLaMADecodeLayer($\mathbf{X}$, $l$)
\ENDFOR
\ENDIF
\end{algorithmic}
\end{algorithm}

\setlength{\tabcolsep}{7pt}
\begin{table}[ht]
\centering
\begin{tabular}{lcccccc}
\toprule
\multicolumn{6}{c}{LLaVA-1.5-7B($C=1\%$)} \\
\midrule
$S$ & $R$ & FLOPs & Latency & MME & AI2D  \\ 
\midrule
1& 1.75\% & 4.03T & 43.88ms & 1836.47 & 54.63 \\
2& 3.50\% & 4.08T & 41.05ms & 1862.15 & 54.53\\
4& 7.00\% & 4.20T & 40.74ms & 1840.29 & 54.99\\
\rowcolor{violet!10}
7& 12.25\% & 4.37T & 40.12ms & 1866.71 & 55.41 \\
14& 24.50\% & 4.78T & 40.24ms & 1861.87 & 55.41 \\
28& 49.00\% & 5.59T & 42.67ms & 1858.10 & 55.31 \\
\midrule
\multicolumn{6}{c}{LLaVA-1.5-7B($C=8\%$)} \\
\midrule
$S$ & $R$ & FLOPs & Latency & MME & AI2D  \\ 
\midrule
1& 1.50\% & 4.38T & 45.84ms & 1841.09 & 55.20 \\
2& 3.00\% & 4.42T & 42.32ms & 1852.64 & 54.70\\
4& 6.00\% & 4.51T & 41.95ms & 1853.37 & 55.18\\
7& 10.50\% & 4.65T & 41.20ms & 1874.07 & 55.38\\
14& 21.00\% & 4.98T & 40.44ms & 1855.12 & 55.41\\
28& 42.00\% & 5.63T & 42.54ms & 1859.85 & 55.38\\
\midrule
\multicolumn{6}{c}{LLaVA-1.5-7B($C=15\%$)} \\
\midrule
$S$ & $R$ & FLOPs & Latency & MME & AI2D  \\ 
\midrule
1& 1.25\% & 4.55T & 45.71ms & 1862.01 & 54.92 \\
2& 2.50\% & 4.59T & 43.33ms & 1869.80 & 54.89\\
4& 5.00\% & 4.67T & 42.03ms & 1863.48 & 55.41\\
7& 8.75\% & 4.76T & 42.37ms & 1867.44 & 55.31\\
14& 17.50\% & 5.08T & 42.74ms & 1855.12 & 55.34\\
28& 35.00\% & 5.65T & 42.20ms & 1863.85 & 55.28\\
\midrule
\multicolumn{6}{c}{LLaVA-1.5-7B($C=22\%$)} \\
\midrule
$S$ & $R$ & FLOPs & Latency & MME & AI2D  \\ 
\midrule
1& 1.00\% & 4.73T & 47.55ms & 1871.05 & 55.38 \\
2& 2.00\% & 4.76T & 44.52ms & 1866.91 & 55.02\\
4& 4.00\% & 4.83T & 43.13ms & 1855.84 & 55.08\\
7& 7.00\% & 4.93T & 42.27ms & 1873.49 & 55.34\\
14& 14.00\% & 5.18T & 42.51ms & 1869.37 & 55.41\\
28& 28.00\% & 5.67T & 42.96ms & 1875.87 & 55.31\\
\midrule
\multicolumn{6}{c}{LLaVA-1.5-7B($C=29\%$)} \\
\midrule
$S$ & $R$ & FLOPs & Latency & MME & AI2D  \\ 
\midrule
1& 0.75\% & 4.90T & 47.53ms & 1873.11 & 55.21 \\
2& 1.50\% & 4.93T & 44.04ms & 1856.40 & 55.18\\
4& 3.00\% & 4.99T & 42.60ms & 1872.19 & 55.18\\
7& 5.25\% & 5.07T & 42.81ms & 1865.72 & 55.25\\
14& 10.50\% & 5.28T & 42.90ms & 1863.12 & 55.38\\
28& 21.00\% & 5.69T & 42.90ms & 1862.37 & 55.38\\
\bottomrule
\end{tabular}
\caption{Results of the pruning stride ``$S$'' and the pruning ratio ``$R$'' per step on LLaVA-1.5-7B.}
\vspace{-0.5cm}
\label{table:a1}
\end{table}

\setlength{\tabcolsep}{7pt}
\begin{table}[ht]
\centering
\begin{tabular}{lcccccc}
\toprule
\multicolumn{6}{c}{LLaVA-1.5-13B($C=5\%$)} \\
\midrule
$S$ & $R$ & FLOPs & Latency & MME & AI2D  \\ 
\midrule
1 & 1.25\% & 7.64T & 68.28ms & 1826.94 & 58.78 \\
2 & 2.50\% & 7.72T & 63.96ms & 1851.51 & 58.91 \\
3 & 3.75\% & 7.80T & 64.54ms & 1833.87 & 59.10 \\
\rowcolor{violet!10}
4 & 5.00\% & 7.88T & 63.15ms & 1830.51 & 58.81 \\
6 & 7.50\% & 8.04T & 63.34ms & 1827.58 & 59.07 \\
9 & 11.25\% & 8.29T & 64.06ms & 1832.09 & 58.71 \\
\midrule
\multicolumn{6}{c}{LLaVA-1.5-13B($C=14\%$)} \\
\midrule
$S$ & $R$ & FLOPs & Latency & MME & AI2D  \\ 
\midrule
1 & 1.00\% & 8.08T & 69.99ms & 1838.19 & 58.81\\
2 & 2.00\% & 8.15T & 66.22ms & 1832.35 & 58.65 \\
3 & 3.00\% & 8.22T & 65.96ms & 1837.00 & 58.81 \\
4 & 4.00\% & 8.29T & 64.63ms & 1826.89 & 58.68\\
6 & 6.00\% & 8.43T & 65.40ms & 1839.08 & 59.13 \\
9 & 9.00\% & 8.64T & 66.40ms & 1835.70 & 58.71 \\
\midrule
\multicolumn{6}{c}{LLaVA-1.5-13B($C=23\%$)} \\
\midrule
$S$ & $R$ & FLOPs & Latency & MME & AI2D  \\ 
\midrule
1 & 0.75\% & 8.52T & 72.15ms & 1839.05 & 58.61\\
2 & 1.50\% & 8.58T & 67.76ms & 1846.45 & 58.71 \\
3 & 2.25\% & 8.63T & 68.17ms & 1838.58 & 58.74 \\
4 & 3.00\% & 8.69T & 67.04ms & 1830.77 & 58.61\\
6 & 4.50\% & 8.81T & 66.84ms & 1844.67 & 59.00 \\
9 & 6.75\% & 8.98T & 67.72ms & 1851.09 & 58.74 \\
\midrule
\multicolumn{6}{c}{LLaVA-1.5-13B($C=32\%$)} \\
\midrule
$S$ & $R$ & FLOPs & Latency & MME & AI2D  \\ 
\midrule
1 & 0.50\% & 9.40T & 78.30ms & 1832.83 & 58.84\\
2 & 1.00\% & 9.44T & 74.24ms & 1853.45 & 58.81 \\
3 & 1.50\% & 9.47T & 73.37ms & 1833.76 & 58.81 \\
4 & 2.00\% & 9.51T & 71.72ms & 1848.60 & 58.78\\
6 & 3.00\% & 9.58T & 71.35ms & 1849.84 & 58.91 \\
9 & 4.50\% & 9.68T & 72.02ms & 1844.09 & 58.81 \\
\midrule
\multicolumn{6}{c}{LLaVA-1.5-13B($C=41\%$)} \\
\midrule
$S$ & $R$ & FLOPs & Latency & MME & AI2D  \\ 
\midrule
1 & 0.25\% & 9.84T & 78.47ms & 1846.59 & 58.94 \\
2 & 0.50\% & 9.87T & 74.35ms & 1858.33 & 58.97 \\
3 & 0.75\% & 9.89T & 73.02ms & 1851.34 & 58.97 \\
4 & 1.00\% & 9.91T & 72.92ms & 1847.56 & 58.91 \\
6 & 1.50\% & 9.96T & 72.09ms & 1845.34 & 58.94 \\
9 & 2.25\% & 10.03T & 71.55ms & 1852.34 & 58.91 \\
\bottomrule
\end{tabular}
\caption{Results of the pruning stride ``$S$'' and the pruning ratio ``$R$'' per step on LLaVA-1.5-13B.}
\vspace{-0.5cm}
\label{table:a2}
\end{table}

\setlength{\tabcolsep}{12pt}
\begin{table*}[t]
\centering
\begin{tabular}{lccccccc}
\toprule
Methods & FLOPs$\downarrow$ & Latency$\downarrow$ & MME$\uparrow$ & AI2D$\uparrow$ & SQA$\uparrow$ & MMMU$\uparrow$ & POPE$\uparrow$  \\ 
\midrule
LLaVA-1.5-7B & 9.38T & 70.80ms & 1861.97 & 55.25 & 69.46 & 35.20 & 86.99 \\
4~$\rightarrow$ 5$\sim$9 & 9.14T & 63.39ms & 1295.72 & 26.59 & 48.29 & 27.90 & 86.04 \\
19~$\rightarrow$ 20$\sim$24 & 9.14T & 63.39ms & 1857.90 & 54.60 & 69.71 & 35.10 & 86.74 \\ 
19~$\rightarrow$ 20$\sim$30 & 8.85T & 59.40ms & 1827.69 & 53.92 & 69.31 & 34.30 & 85.71 \\
\midrule
LLaVA-1.5-13B & 17.81T & 128.49ms & 1817.95 & 59.26 & 72.78 & 35.00 & 87.09 \\
5~$\rightarrow$ 6$\sim$10 & 17.44T & 117.96ms & 1432.31 & 26.98 & 56.02 & 28.30 & 86.56 \\
26~$\rightarrow$ 27$\sim$31 & 17.44T & 117.96ms & 1811.63 & 59.23 & 72.68 & 34.70 & 87.18 \\ 
26~$\rightarrow$ 27$\sim$37 & 17.00T & 111.63ms & 1809.59 & 59.29 & 72.58 & 34.70 & 87.33 \\
\bottomrule
\end{tabular}
\caption{Results of implementation QK heredity across multiple layers. ``4~$\rightarrow$ 5$\sim$9'' means reusing attention score of layer4 from layer5 to layer9.}
\label{table:qk_dataset_multi}
\end{table*}

\setlength{\tabcolsep}{12pt}
\begin{table*}[t]
\centering
\begin{tabular}{lcccccc}
\toprule
Methods & MME$\uparrow$ & AI2D$\uparrow$ & SQA$\uparrow$ & MMMU$\uparrow$ & POPE$\uparrow$ & Coco2017\_cap\_val$\uparrow$ \\ 
\midrule
LLaVA-1.5-7B & 1861.97 & 55.25 & 69.46 & 35.20 & 86.99 & 110.43\\
1~$\rightarrow$ 2 & 1508.03 & 10.98 & 7.29 & 27.00 & 85.38 & 98.79\\
2~$\rightarrow$ 3 & 945.18 & 14.86 & 10.51 & 22.90 & 75.50 & 93.62\\
3~$\rightarrow$ 4 & 1812.92 & 52.49 & 66.09 & 33.60 & 86.90 & 109.04\\
4~$\rightarrow$ 5 & 1767.67 & 52.91 & 66.83 & 33.20 & 86.49 & 111.24\\
5~$\rightarrow$ 6 & 1787.80 & 52.82 & 65.94 & 32.60 & 86.88 & 111.16\\
6~$\rightarrow$ 7 & 1793.07 & 49.19 & 66.53 & 33.20 & 80.82 & 110.73\\
7~$\rightarrow$ 8 & 1780.78 & 53.63 & 68.47 & 35.40 & 87.21 & 110.80\\
8~$\rightarrow$ 9 & 1660.70 & 51.78 & 65.05 & 34.00 & 86.62 & 109.46\\
9~$\rightarrow$ 10 & 1640.23 & 51.36 & 67.72 & 34.70 & 87.56 & 109.63\\
10~$\rightarrow$ 11 & 1616.02 & 49.48 & 64.05 & 32.10 & 86.28 & 112.36\\
11~$\rightarrow$ 12 & 1722.73 & 50.42 & 65.49 & 33.10 & 80.66 & 111.44\\
12~$\rightarrow$ 13 & 1769.66 & 51.81 & 66.19 & 33.80 & 87.04 & 107.43\\
13~$\rightarrow$ 14 & 1695.21 & 52.46 & 65.64 & 36.20 & 87.80 & 109.84\\
14~$\rightarrow$ 15 & 1866.85 & 53.24 & 67.97 & 33.40 & 85.19 & 110.18\\
15~$\rightarrow$ 16 & 1879.09 & 54.53 & 68.96 & 36.00 & 86.78 & 105.25\\
16~$\rightarrow$ 17 & 1676.24 & 53.98 & 69.31 & 34.30 & 87.39 & 113.03\\
17~$\rightarrow$ 18 & 1876.72 & 54.63 & 69.41 & 35.60 & 86.52 & 108.81\\
18~$\rightarrow$ 19 & 1854.47 & 54.50 & 67.87 & 36.00 & 86.59 & 102.96\\
19~$\rightarrow$ 20 & 1869.39 & 54.86 & 69.16 & 34.80 & 86.98 & 110.04\\
20~$\rightarrow$ 21 & 1880.37 & 55.34 & 69.41 & 35.80 & 86.91 & 109.73\\
21~$\rightarrow$ 22 & 1858.02 & 55.12 & 69.11 & 35.20 & 87.07 & 109.37\\
22~$\rightarrow$ 23 & 1864.37 & 55.12 & 69.56 & 35.70 & 86.99 & 109.82\\
23~$\rightarrow$ 24 & 1847.22 & 55.21 & 69.41 & 35.80 & 86.89 & 110.00\\
24~$\rightarrow$ 25 & 1855.57 & 55.12 & 69.26 & 35.80 & 86.91 & 110.45\\
25~$\rightarrow$ 26 & 1874.14 & 54.99 & 69.36 & 35.40 & 86.82 & 110.54\\
26~$\rightarrow$ 27 & 1877.16 & 55.15 & 69.46 & 35.40 & 86.46 & 109.69\\
27~$\rightarrow$ 28 & 1855.31 & 55.08 & 69.36 & 35.00 & 86.94 & 110.00\\
28~$\rightarrow$ 29 & 1869.93 & 55.21 & 69.61 & 35.40 & 86.89 & 109.17\\
29~$\rightarrow$ 30 & 1862.70 & 55.25 & 69.26 & 35.00 & 86.93 & 110.15\\
30~$\rightarrow$ 31 & 1837.34 & 53.95 & 69.21 & 34.60 & 86.74 & 101.63\\
31~$\rightarrow$ 32 & 1640.46 & 54.47 & 69.45 & 34.10 & 84.15 & 87.10\\
\bottomrule
\end{tabular}
\caption{Results of the implementation QK heredity ($n=1$) on each layer in LLaVA-1.5-7B. ``0~$\rightarrow$ 1'' means reusing attention score of layer1 in layer2.}
\label{table:qk_heredity_datasets7b}
\end{table*}

\setlength{\tabcolsep}{12pt}
\begin{table*}[t]
\centering
\begin{tabular}{lcccccc}
\toprule
Methods & MME$\uparrow$ & AI2D$\uparrow$ & SQA$\uparrow$ & MMMU$\uparrow$ & POPE$\uparrow$ & Coco2017\_cap\_val$\uparrow$ \\ 
\midrule
LLaVA-1.5-13B & 1817.95 & 59.26 & 72.78 & 35.20 & 87.09 & 115.57\\
1~$\rightarrow$ 2 & 1274.76 & 0.03 & 0.19 & 23.40 & 80.22 & 70.07\\
2~$\rightarrow$ 3 & 1678.44 & 50.62 & 66.39 & 31.10 & 87.49 & 118.45\\
3~$\rightarrow$ 4 & 1780.24 & 55.93 & 69.21 & 34.20 & 88.04 & 116.08\\
4~$\rightarrow$ 5 & 1097.25 & 23.51 & 22.76 & 27.10 & 83.69 & 116.18\\
5~$\rightarrow$ 6 & 1826.21 & 57.67 & 71.10 & 34.80 & 87.01 & 114.25\\
6~$\rightarrow$ 7 & 1813.88 & 58.26 & 70.40 & 34.30 & 87.63 & 115.56\\
7~$\rightarrow$ 8 & 1805.96 & 57.90 & 72.58 & 34.70 & 86.77 & 113.91\\
8~$\rightarrow$ 9 & 1831.15 & 57.12 & 71.89 & 34.40 & 87.29 & 116.16\\
9~$\rightarrow$ 10 & 1719.02 & 56.44 & 70.60 & 34.80 & 87.92 & 114.62\\
10~$\rightarrow$ 11 & 1751.42 & 55.44 & 71.29 & 33.40 & 88.24 & 115.00\\
11~$\rightarrow$ 12 & 1729.24 & 54.05 & 71.59 & 33.60 & 82.16 & 117.85\\
12~$\rightarrow$ 13 & 1726.85 & 57.51 & 70.65 & 35.90 & 87.81 & 114.11\\
13~$\rightarrow$ 14 & 1750.54 & 58.52 & 71.94 & 33.60 & 86.87 & 117.13\\
14~$\rightarrow$ 15 & 1784.11 & 59.03 & 72.38 & 35.10 & 88.32 & 115.36\\
15~$\rightarrow$ 16 & 1806.68 & 57.97 & 71.34 & 34.40 & 86.93 & 115.82\\
16~$\rightarrow$ 17 & 1819.35 & 58.42 & 71.69 & 35.30 & 87.32 & 115.55\\
17~$\rightarrow$ 18 & 1827.88 & 58.68 & 72.43 & 35.60 & 87.43 & 117.03\\
18~$\rightarrow$ 19 & 1793.94 & 58.84 & 72.53 & 35.40 & 87.50 & 114.64\\
19~$\rightarrow$ 20 & 1820.02 & 58.81 & 72.83 & 34.70 & 87.22 & 115.49\\
20~$\rightarrow$ 21 & 1844.59 & 59.46 & 72.63 & 34.70 & 87.13 & 117.55\\
21~$\rightarrow$ 22 & 1832.85 & 58.94 & 72.68 & 35.10 & 87.21 & 112.49\\
22~$\rightarrow$ 23 & 1824.77 & 59.29 & 72.58 & 35.10 & 87.22 & 114.96\\
23~$\rightarrow$ 24 & 1834.74 & 59.16 & 72.88 & 35.10 & 87.13 & 115.43\\
24~$\rightarrow$ 25 & 1821.03 & 59.10 & 72.78 & 35.00 & 87.17 & 114.74\\
25~$\rightarrow$ 26 & 1839.22 & 59.26 & 72.83 & 34.80 & 87.18 & 115.79\\
26~$\rightarrow$ 27 & 1828.97 & 59.42 & 72.83 & 35.00 & 87.13 & 116.09\\
27~$\rightarrow$ 28 & 1817.33 & 59.20 & 72.78 & 35.00 & 87.19 & 114.75\\
28~$\rightarrow$ 29 & 1814.94 & 59.29 & 72.78 & 34.80 & 87.21 & 115.76\\
29~$\rightarrow$ 30 & 1821.73 & 59.29 & 72.83 & 34.90 & 87.14 & 115.78\\
30~$\rightarrow$ 31 & 1823.64 & 59.33 & 72.83 & 35.10 & 87.10 & 114.65\\
31~$\rightarrow$ 32 & 1827.74 & 59.36 & 72.78 & 35.00 & 87.14 & 115.09\\
32~$\rightarrow$ 33 & 1822.36 & 59.29 & 72.78 & 35.10 & 87.11 & 114.45\\
33~$\rightarrow$ 34 & 1828.29 & 59.13 & 72.88 & 35.10 & 87.12 & 116.01\\
34~$\rightarrow$ 35 & 1825.93 & 59.23 & 72.78 & 35.00 & 87.11 & 116.02\\
35~$\rightarrow$ 36 & 1820.26 & 59.33 & 72.88 & 35.10 & 87.09 & 115.19\\
36~$\rightarrow$ 37 & 1829.73 & 59.26 & 72.88 & 34.80 & 87.17 & 115.84\\
37~$\rightarrow$ 38 & 1811.64 & 59.29 & 73.03 & 35.10 & 87.17 & 112.89\\
38~$\rightarrow$ 39 & 1825.73 & 59.26 & 72.83 & 34.90 & 87.37 & 114.05\\
39~$\rightarrow$ 40 & 1821.00 & 46.76 & 63.41 & 34.40 & 22.00 & 104.69\\
\bottomrule
\end{tabular}
\caption{Results of the implementation QK heredity ($n=1$) on each layer in LLaVA-1.5-13B. ``0~$\rightarrow$ 1'' means reusing attention score of layer1 in layer2.}
\label{table:qk_heredity_datasets13b}
\end{table*}

\end{document}